\pdfoutput=1

\documentclass[11pt]{article}

\usepackage{acl}

\usepackage{times}
\usepackage{latexsym}

\usepackage[T1]{fontenc}

\usepackage{microtype}
\usepackage{inconsolata}

\usepackage{graphicx}

\usepackage{array}
\usepackage{here}
\usepackage{float}
\usepackage{multicol}

\usepackage{multirow,tabularx}
\usepackage{array}
\usepackage{here}
\usepackage{float}
\usepackage{multicol}
\usepackage{arydshln}

\usepackage{amsmath}
\usepackage{amssymb}
\usepackage{amsfonts}
\usepackage{latexsym}
\usepackage{comment}
\usepackage{color}
\usepackage{seqsplit}

\newcommand{\tabH}{\rule{0pt}{2.2ex}}
\newcommand{\bhline}{\noalign{\hrule height 1.2pt}}
\newcommand{\footlink}[1]{\footnote{\url{#1}}}
\newcommand{\foothref}[2]{\footnote{\href{#1}{\texttt{\seqsplit{#2}}}}}
\newcommand{\foothreftext}[2]{\footnotetext{\href{#1}{\texttt{\seqsplit{#2}}}}}

\newcommand{\footsep}{{$^{,}$}}

\newcommand{\supplement}[1]{{\fontsize{10}{10} \texttt{\seqsplit{(#1)}}} }

\title{Japanese SimCSE Technical Report}

\author{
  Hayato Tsukagoshi \hspace{2em} Ryohei Sasano \hspace{2em} Koichi Takeda \\
  Graduate School of Informatics, Nagoya University\\
  \texttt{tsukagoshi.hayato.r2@s.mail.nagoya-u.ac.jp}, \\
  \texttt{\{sasano,takedasu\}@i.nagoya-u.ac.jp} \\
}

\begin{document}
\maketitle
\begin{abstract}
We report the development of Japanese SimCSE, Japanese sentence embedding models fine-tuned with SimCSE.
Since there is a lack of sentence embedding models for Japanese that can be used as a baseline in sentence embedding research, we conducted extensive experiments on Japanese sentence embeddings involving 24 pre-trained Japanese or multilingual language models, five supervised datasets, and four unsupervised datasets.
In this report, we provide the detailed training setup for Japanese SimCSE and their evaluation results.
\end{abstract}

\section{Introduction}

Sentence embeddings provide dense vector representations of natural language sentences and have gained traction in tasks such as retrieval, question answering, and, more recently, Retrieval Augmented Generation (RAG).
Although various methods exist to produce sentence embeddings, recent approaches have shown promise by fine-tuning pre-trained language models using contrastive learning.
Among them, SimCSE~\cite{SimCSE} is a pioneering work of contrastive sentence embeddings and offers techniques for both unsupervised and supervised settings.
In the unsupervised setting, SimCSE leverages Dropout in pre-trained language models as a data augmentation technique, processing the same sentence twice through the model. It then treats pairs of embeddings from the same sentence as positive samples for contrastive learning. In the supervised approach, it utilizes the Natural Language Inference (NLI) dataset, such as the Stanford NLI (SNLI) dataset~\cite{SNLI} and the Multi-Genre NLI (MNLI) dataset~\cite{MNLI}, to treat semantically similar sentences as positive samples for contrastive learning.
Additionally, to further emphasize generating embeddings that capture differences in meaning, SimCSE uses sentence pairs labeled as contradictions in the NLI dataset as hard negatives.

SimCSE has arguably become the de facto standard for sentence embeddings, demonstrating wide-ranging and impressive performance and leading to numerous derivative studies~\cite{DiffCSE,PromptBERT,PromptEOL}.
However, many of these studies focus on English, with a lack of comprehensive research on Japanese sentence embeddings.

In this report, we present extensive experiments using various Japanese or multilingual pre-trained language models, training datasets, and hyperparameters to perform a thorough evaluation of Japanese SimCSE.
Additionally, we release four pre-trained Japanese sentence embedding models fine-tuned under promising settings and present their evaluation results to encourage further research.
Our models, detailed results, and codebases are publicly available\footlink{https://github.com/hppRC/simple-simcse-ja}.

\section{Training Datasets}

First, we conducted comprehensive experiments to investigate promising configurations for training Japanese sentence embedding models.
As SimCSE offers both supervised and unsupervised settings, we will discuss training datasets for each in detail.

\subsection{Datasets for Supervised SimCSE}

To obtain a Japanese Supervised SimCSE model, Japanese NLI datasets are required.
However, it remains uncertain which dataset should be used to train the Japanese SimCSE model.
To ensure fair comparisons between datasets, we standardized the format for each training dataset individually and subsequently applied a common text preprocessing procedure to all datasets\footlink{https://github.com/hppRC/simple-simcse-ja/blob/main/src/datasets/common.py} using Konoha\footlink{https://github.com/himkt/konoha}.
Therefore, we conducted experiments using the following five datasets.

\begin{description}
\item[JSNLI\protect\footnotemark]\foothreftext{https://nlp.ist.i.kyoto-u.ac.jp/?\%E6\%97\%A5\%E6\%9C\%AC\%E8\%AA\%9ESNLI\%28JSNLI\%29\%E3\%83\%87\%E3\%83\%BC\%E3\%82\%BF\%E3\%82\%BB\%E3\%83\%83\%E3\%83\%88}{https://nlp.ist.i.kyoto-u.ac.jp}{
is a Japanese translation of SNLI, a standard benchmark for NLI.
Machine translation was applied to SNLI, and then it was refined by applying filtering through crowdsourcing for evaluation data and automated filtering for training data.
}

\item[JaNLI]{\cite{JaNLI}
is a synthetically generated Japanese Adversarial NLI dataset designed to assess understanding of Japanese linguistic phenomena and to illuminate the vulnerabilities of models.
}

\item[NU-NLI]{
is our new collection of NLI datasets, specifically \textbf{NU-SNLI}, \textbf{NU-MNLI}, and \textbf{NU-SNLI+MNLI} or simply \textbf{NU-NLI}, derived from translating the Stanford NLI (SNLI) and Multi-Genre NLI (MNLI) datasets into Japanese using ChatGPT\foothref{https://platform.openai.com/docs/models/gpt-3-5}{https://platform.openai.com}\supplement{gpt-3.5-turbo-0301}.
While these datasets were created for this report, and we initially planned to make them available as a consolidated NU-NLI dataset, their public release has been postponed due to licensing concerns.
}
\end{description}

\subsection{Datasets for Unsupervised SimCSE}

Similarly to Supervised SimCSE, to observe performance differences due to the nature of unsupervised datasets, we used the four unsupervised datasets for fine-tuning with Unsupervised SimCSE.
The same common text preprocessing as for Supervised SimCSE was applied.

\begin{description}
\item[Wiki40B]{\cite{Wiki40B}
is a multilingual language model benchmark dataset that is composed of 40+ languages spanning several scripts and linguistic families. 
Wiki40B is a meticulously preprocessed Wikipedia dataset, and its quality as unsupervised text is relatively high.
}

\item[Wikipedia\protect\footnotemark]\footnotetext{\url{https://huggingface.co/datasets/wikipedia}}{
is a dataset preprocessed from the dump of Japanese Wikipedia articles as of January 1, 2023.
The preprocessing method was inspired by the process used when creating the pre-training corpus for Tohoku University's Japanese BERT\protect\footnotemark.
}
\footnotetext{
\href{https://github.com/cl-tohoku/bert-japanese/blob/041b855db36942d42f2ca062da14416eaef93223/make_corpus_wiki.py\#L54}{\texttt{https://github.com/cl-tohoku/bert-japanese}}
}

\item[BCCWJ\protect\footnotemark]\footnotetext{\url{https://clrd.ninjal.ac.jp/bccwj}}{
(Balanced Corpus of Contemporary Written Japanese) consists of 104.3 million words across genres such as books, magazines, newspapers, white papers, blogs, Internet forums, textbooks, and law.
Each sample is drawn for each genre randomly.
}

\item[CC100\protect\footnotemark]\footnotetext{\url{https://data.statmt.org/cc-100/}}{\cite{CC100-1,CC100-2}
is a collection of monolingual large-scale web corpora constructed from January-December 2018 Commoncrawl snapshots.
}

\end{description}

\section{Evaluation}

We fine-tuned various models using SimCSE across diverse datasets and hyperparameters, and then evaluated them on the Semantic Textual Similarity (STS) tasks.

\subsection{Settings}

For evaluating sentence embedding models, the STS tasks have often been employed.
The STS task evaluates the efficacy of sentence embeddings by measuring the correlation between human-annotated similarity scores for a pair of sentences and the computed semantic similarity of the embeddings of those sentences.
Cosine similarity is commonly employed to measure the similarity between sentence embeddings.

In terms of English datasets, STS12--16~\cite{STS12,STS13,STS14,STS15,STS16}, STS Benchmark~\cite{STSB}, and SICK~\cite{SICK} are commonly used for benchmarking sentence embeddings.
In recent years, there has been a growing effort by researchers to develop Japanese STS datasets.
In this report, we utilize these Japanese STS datasets for evaluations based on the STS task.
Specifically, we used JSICK~\cite{JSICK} and JSTS from JGLUE~\cite{JGLUE}.
\begin{description}

\item[JSICK]{
is a Japanese NLI and STS dataset derived from manually translating the English SICK dataset~\cite{SICK} into Japanese.
JSICK comprises both a validation set and a test set.
There also exists another split known as the JSICK-stress Test Set, though we did not use it in this report.
}

\item[JSTS]{
is a Japanese STS dataset constructed from sentences in the Japanese version of the MS COCO Caption Dataset, the YJ Captions Dataset~\cite{YJCaptions}.
Currently, only the train set and validation set of JSTS are publicly available.
In this report, we employed both the train set and the validation set as evaluation datasets.
}
\end{description}

It is worth noting that JSTS shares some of its data with JGLUE's JNLI.
While not experimented in this report, if one were to use JSTS as a development set, there could be a potential data leakage when training models on JNLI.
This might result in artificially high development scores and possibly overfitting the models.
Hence, for this research, we opted to use the validation set of JSICK as our development set.

For the evaluation score, following standard procedures in evaluating sentence embeddings, we used the Spearman's rank correlation coefficient between the model's similarity scores and the human-annotated similarity scores.
We employed the cosine similarity for calculating similarities between sentence embeddings.

\paragraph{Training Details}
We fine-tuned models with both Supervised SimCSE and Unsupervised SimCSE.
For the technical details of SimCSE, please refer to the original paper.
Experiments were conducted with 24 Japanese or multilingual BERT~\cite{BERT}-like pre-trained models, combinations of 5 supervised and 4 unsupervised datasets, 4 batch sizes \{64, 128, 256, 512\}, and 3 learning rates \{1e-5, 3e-5, 5e-5\}.
We used the AdamW~\cite{AdamW} optimizer. 
All experiments were conducted using NVIDIA A100 and NVIDIA RTX A6000 GPUs.
For training, we employed BF16 as the data type and utilized gradient checkpointing\footlink{https://github.com/cybertronai/gradient-checkpointing} to reduce memory consumption.
From the results of preliminary experiments, the performance difference between this method and conventional training with FP32 was negligible.
Among the models used in our experiments, some require tokenization by specific tokenizers (e.g., Juman++\footlink{https://github.com/ku-nlp/jumanpp}~\cite{Juman}) prior to training (e.g., Waseda University's RoBERTa\footlink{https://huggingface.co/nlp-waseda/roberta-base-japanese}).
For these models, we applied the appropriate tokenization using corresponding tokenizers.

We employed a linear warmup for the learning rate scheduling.
The learning rate was increased linearly for the first 10\% of the total training steps and then decreased linearly thereafter.
While the original SimCSE paper set the maximum sequence length to 32, we speculated that the dynamics might differ for Japanese, prompting us to use a longer value.
Therefore, we set the maximum sequence length to 64.
In general, the temperature parameter used for similarity scaling in contrastive learning has a significant impact on performance.
Therefore, we conducted temperature parameter tuning using Optuna~\cite{Optuna} in our preliminary experiments.
As a result, in our setting, we confirmed that as long as the temperature parameter is set to a typical value of around 0.05 and both the batch size and learning rate are appropriately tuned, the performance remains unaffected by the temperature parameter.
Consequently, we adopted the default temperature parameter value of 0.05.
During training, performance on a development set is evaluated at some intervals, and the checkpoint with the highest evaluation score on the development set is used for the final evaluation.
For English models, the dev set from the STS Benchmark is utilized as the development set.
In our case, we employed JSICK (train) as our development set.

\paragraph{Evaluation Steps}
In this report, we conducted experiments using multiple datasets of varying sizes.
To mitigate the potential impact of dataset size on performance, we standardized the number of training examples to $2^{20}$ for all datasets by random sampling\footnote{
The reason for selecting $2^{20} \approx$ 1M as the number of training examples is that Unsupervised SimCSE is trained with one million sentences.
}.
During fine-tuning with SimCSE, \citet{SimCSE} evaluated the model on the development set every 250 steps.
However, when fixing the number of training examples, the number of training steps varies depending on the batch size.
This could lead to smaller batch sizes resulting in more frequent evaluations, potentially skewing comparisons.
Therefore, in this report, we fixed the number of evaluations to $2^6$, ensuring evaluations occurred at varied intervals depending on the batch size.
For instance, when setting the batch size to $2^8 = 256$, the model is evaluated on the development set every $2^{20} \div 2^8 \div 2^6 = 64$ steps.

\paragraph{Experimental Stability}
Based on our insights from the replication study we conducted on English SimCSE\footlink{https://github.com/hppRC/simple-simcse}, the performance of SimCSE can vary depending on the random seed value and hyperparameters.
To alleviate the variability, we averaged evaluation scores across five runs.
As a result, we carried out experiments 7,200 times in the supervised setting and 5,760 times in the unsupervised setting.

\subsection{Results}

\begin{table*}[t!]
\small
\renewcommand{\arraystretch}{1.0}
\tabcolsep 4pt
\centering
\begin{tabular}{@{\hspace{2ex}}lc@{\hspace{3ex}}ccc@{\hspace{4ex}}c@{\hspace{1ex}}}
\bhline
\tabH \hspace{-1ex}Model Id & JSICK (val) & JSICK (test) & JSTS (train) & JSTS (val) & Avg. \\
\bhline
\multicolumn{6}{l}{\tabH Base Size} \\
\bhline
\tabH \href{https://huggingface.co/cl-tohoku/bert-base-japanese-v3}{cl-tohoku/bert-base-japanese-v3} & 83.60 & 82.66 & \textbf{77.34} & \textbf{80.70} & 80.23 \\
\href{https://huggingface.co/cl-tohoku/bert-base-japanese-v2}{cl-tohoku/bert-base-japanese-v2} & 84.20 & \textbf{83.39} & 77.03 & \textbf{80.70} & \textbf{80.37} \\
\href{https://huggingface.co/cl-tohoku/bert-base-japanese}{cl-tohoku/bert-base-japanese} & 83.39 & 82.44 & 75.25 & 78.46 & 78.72 \\
\href{https://huggingface.co/cl-tohoku/bert-base-japanese-whole-word-masking}{cl-tohoku/bert-base-japanese-whole-word-masking} & 83.29 & 82.32 & 75.79 & 79.01 & 79.04 \\
\href{https://huggingface.co/studio-ousia/luke-japanese-base-lite}{studio-ousia/luke-japanese-base-lite} & 82.89 & 81.72 & 75.64 & 79.34 & 
78.90 \\
\hline
\tabH \href{https://huggingface.co/ku-nlp/deberta-v2-base-japanese}{ku-nlp/deberta-v2-base-japanese} & 81.90 & 80.78 & 74.71 & 78.39 & 77.96 \\
\href{https://huggingface.co/nlp-waseda/roberta-base-japanese}{nlp-waseda/roberta-base-japanese} & 82.94 & 82.00 & 75.65 & 79.63 & 79.09 \\
\href{https://huggingface.co/megagonlabs/roberta-long-japanese}{megagonlabs/roberta-long-japanese} & 82.25 & 80.77 & 72.39 & 76.54 & 76.57 \\
\hline
\tabH \href{https://huggingface.co/cl-tohoku/bert-base-japanese-char-v3}{cl-tohoku/bert-base-japanese-char-v3} & 82.57 & 81.35 & 75.75 & 78.62 & 78.57 \\
\href{https://huggingface.co/cl-tohoku/bert-base-japanese-char-v2}{cl-tohoku/bert-base-japanese-char-v2} & 83.38 & 81.95 & 74.98 & 78.64 & 78.52 \\
\href{https://huggingface.co/cl-tohoku/bert-base-japanese-char}{cl-tohoku/bert-base-japanese-char} & 82.89 & 81.40 & 74.35 & 77.79 & 77.85 \\
\href{https://huggingface.co/ku-nlp/roberta-base-japanese-char-wwm}{ku-nlp/roberta-base-japanese-char-wwm} & 82.80 & 80.62 & 74.35 & 78.54 & 77.84 \\
\hline
\tabH \href{https://huggingface.co/bert-base-multilingual-cased}{bert-base-multilingual-cased} & 83.46 & 82.12 & 73.33 & 76.82 & 77.42 \\
\href{https://huggingface.co/xlm-roberta-base}{xlm-roberta-base} & 80.29 & 78.42 & 72.54 & 76.02 & 75.66 \\
\href{https://huggingface.co/studio-ousia/mluke-base-lite}{studio-ousia/mluke-base-lite} & 83.48 & 81.96 & 74.97 & 78.47 & 78.47 \\
\bhline
\multicolumn{6}{l}{\tabH Large Size} \\
\bhline
\tabH \href{https://huggingface.co/cl-tohoku/bert-large-japanese-v2}{cl-tohoku/bert-large-japanese-v2} & 83.97 & 82.63 & \textbf{79.44} & \textbf{82.98} & \textbf{81.68} \\
\href{https://huggingface.co/cl-tohoku/bert-large-japanese}{cl-tohoku/bert-large-japanese} & 83.70 & 82.54 & 76.49 & 80.09 & 79.71 \\
\href{https://huggingface.co/studio-ousia/luke-japanese-large-lite}{studio-ousia/luke-japanese-large-lite} & 83.82 & 82.50 & 78.94 & 82.24 & 81.23 \\
\hline
\tabH \href{https://huggingface.co/nlp-waseda/roberta-large-japanese}{nlp-waseda/roberta-large-japanese} & 84.42 & \textbf{83.08} & 79.28 & 82.63 & 81.66 \\
\href{https://huggingface.co/ku-nlp/deberta-v2-large-japanese}{ku-nlp/deberta-v2-large-japanese} & 79.81 & 79.47 & 77.32 & 80.29 & 79.03 \\
\hline
\tabH \href{https://huggingface.co/cl-tohoku/bert-large-japanese-char-v2}{cl-tohoku/bert-large-japanese-char-v2} & 83.63 & 82.14 & 77.97 & 80.88 & 80.33 \\
\href{https://huggingface.co/ku-nlp/roberta-large-japanese-char-wwm}{ku-nlp/roberta-large-japanese-char-wwm} & 83.30 & 81.87 & 77.54 & 80.90 & 80.10 \\
\hline
\tabH \href{https://huggingface.co/xlm-roberta-large}{xlm-roberta-large} & 83.59 & 82.04 & 76.63 & 79.91 & 79.53 \\
\href{https://huggingface.co/studio-ousia/mluke-large-lite}{studio-ousia/mluke-large-lite} & 84.02 & 82.34 & 77.69 & 80.01 & 80.01 \\
\bhline
\end{tabular}
\vspace{-1ex}
\caption{
Evaluation results for Supervised SimCSE.
Values in the table represent the Spearman's rank correlation coefficient multiplied by 100.
``Model Id'' refers to the identifier of the pre-trained model available on HuggingFace\protect\footnotemark.
The JSICK (val) dataset was used as a development set during training.
``Avg.'' indicates the average performance on three datasets: JSICK (test), JSTS (train), and JSTS (val).
}
\label{tab:main-sup}

\vspace{4ex}

\begin{tabular}{@{\hspace{2ex}}lc@{\hspace{3ex}}ccc@{\hspace{4ex}}c@{\hspace{1ex}}}
\bhline
\tabH \hspace{-1ex}Model Id & JSICK (val) & JSICK (test) & JSTS (train) & JSTS (val) & Avg. \\
\bhline
\multicolumn{6}{l}{\tabH Base Size} \\
\bhline
\tabH \href{https://huggingface.co/cl-tohoku/bert-base-japanese-v3}{cl-tohoku/bert-base-japanese-v3} & 79.17 & 78.47 & \textbf{74.82} & \textbf{78.70} & \textbf{77.33} \\
\href{https://huggingface.co/cl-tohoku/bert-base-japanese-v2}{cl-tohoku/bert-base-japanese-v2} & 80.25 & 79.72 & 72.75 & 77.65 & 76.71 \\
\href{https://huggingface.co/cl-tohoku/bert-base-japanese}{cl-tohoku/bert-base-japanese} & 76.94 & 76.90 & 72.29 & 75.92 & 75.04 \\
\href{https://huggingface.co/cl-tohoku/bert-base-japanese-whole-word-masking}{cl-tohoku/bert-base-japanese-whole-word-masking} & 77.52 & 77.37 & 73.23 & 77.14 & 75.91 \\
\href{https://huggingface.co/studio-ousia/luke-japanese-base-lite}{studio-ousia/luke-japanese-base-lite} & 81.29 & \textbf{80.29} & 72.91 & 78.12 & 77.11 \\
\hline
\tabH \href{https://huggingface.co/ku-nlp/deberta-v2-base-japanese}{ku-nlp/deberta-v2-base-japanese} & 75.51 & 75.23 & 72.07 & 76.54 & 74.61 \\
\href{https://huggingface.co/nlp-waseda/roberta-base-japanese}{nlp-waseda/roberta-base-japanese} & 77.54 & 77.47 & 74.09 & 78.95 & 76.84 \\
\href{https://huggingface.co/megagonlabs/roberta-long-japanese}{megagonlabs/roberta-long-japanese} & 74.53 & 73.95 & 63.10 & 68.72 & 68.59 \\
\hline
\tabH \href{https://huggingface.co/cl-tohoku/bert-base-japanese-char-v3}{cl-tohoku/bert-base-japanese-char-v3} & 78.39 & 78.18 & 73.36 & 77.74 & 76.42 \\
\href{https://huggingface.co/cl-tohoku/bert-base-japanese-char-v2}{cl-tohoku/bert-base-japanese-char-v2} & 79.29 & 79.00 & 71.36 & 75.60 & 75.32 \\
\href{https://huggingface.co/cl-tohoku/bert-base-japanese-char}{cl-tohoku/bert-base-japanese-char} & 77.27 & 76.94 & 69.25 & 73.00 & 73.07 \\
\href{https://huggingface.co/ku-nlp/roberta-base-japanese-char-wwm}{ku-nlp/roberta-base-japanese-char-wwm} & 72.21 & 72.21 & 69.73 & 74.69 & 72.21 \\
\hline
\tabH \href{https://huggingface.co/bert-base-multilingual-cased}{bert-base-multilingual-cased} & 78.45 & 78.23 & 67.60 & 72.36 & 72.73 \\
\href{https://huggingface.co/xlm-roberta-base}{xlm-roberta-base} & 78.70 & 78.37 & 66.63 & 71.28 & 72.09 \\
\href{https://huggingface.co/studio-ousia/mluke-base-lite}{studio-ousia/mluke-base-lite} & 80.38 & 79.83 & 70.79 & 75.31 & 75.31 \\

\bhline
\multicolumn{6}{l}{\tabH Large Size} \\
\bhline

\tabH \href{https://huggingface.co/cl-tohoku/bert-large-japanese-v2}{cl-tohoku/bert-large-japanese-v2} & 79.54 & 79.14 & \textbf{77.18} & 81.00 & 79.11 \\
\href{https://huggingface.co/cl-tohoku/bert-large-japanese}{cl-tohoku/bert-large-japanese} & 78.54 & 78.30 & 72.87 & 76.74 & 75.97 \\
\href{https://huggingface.co/studio-ousia/luke-japanese-large-lite}{studio-ousia/luke-japanese-large-lite} & 79.02 & 78.64 & 75.61 & 79.71 & 77.99 \\
\hline
\tabH \href{https://huggingface.co/nlp-waseda/roberta-large-japanese}{nlp-waseda/roberta-large-japanese} & 82.94 & \textbf{82.56} & 76.04 & \textbf{81.28} & \textbf{79.96} \\
\href{https://huggingface.co/ku-nlp/deberta-v2-large-japanese}{ku-nlp/deberta-v2-large-japanese} & 74.60 & 74.95 & 73.49 & 77.34 & 75.26 \\
\hline
\tabH \href{https://huggingface.co/cl-tohoku/bert-large-japanese-char-v2}{cl-tohoku/bert-large-japanese-char-v2} & 79.07 & 78.73 & 75.68 & 79.10 & 77.83 \\
\bhline
\end{tabular}
\vspace{-1ex}
\caption{
Evaluation results for Unsupervised SimCSE.
Values in the table represent the Spearman's rank correlation coefficient multiplied by 100.
The meaning of each column is the same as in Table~\ref{tab:main-sup}.
}
\label{tab:main-unsup}
\renewcommand{\arraystretch}{1}
\end{table*}
\foothreftext{https://huggingface.co}{https://huggingface.co}

\begin{table}[t!]
\renewcommand{\arraystretch}{1.1}
\small
\tabcolsep 4pt
\centering
\begin{tabular}{lcc}
\bhline
\tabH Model Id & Sup. & Unsup. \\
\bhline
\tabH nlp-waseda/roberta-large-japanese & 1.20 & 1.00 \\
cl-tohoku/bert-large-japanese-v2 & 2.20 & 4.25 \\
studio-ousia/luke-japanese-large-lite & 2.80 & 3.50 \\
cl-tohoku/bert-base-japanese-v3 & 6.40 & 4.50 \\
studio-ousia/mluke-large-lite & 6.60 & 8.00 \\
cl-tohoku/bert-base-japanese-v2 & 6.80 & 10.25 \\
cl-tohoku/bert-large-japanese-char-v2 & 7.00 & 5.00 \\
ku-nlp/roberta-large-japanese-char-wwm & 8.00 & 15.25 \\
cl-tohoku/bert-large-japanese & 9.00 & 10.50 \\
studio-ousia/luke-japanese-base-lite & 10.00 & 7.00 \\
xlm-roberta-large & 11.80 & 5.00 \\
nlp-waseda/roberta-base-japanese & 12.80 & 8.50 \\
ku-nlp/deberta-v2-large-japanese & 13.20 & 13.25 \\
\tabH cl-tohoku/\vspace{-0.2ex} & \multirow{2}{*}{13.80} & \multirow{2}{*}{16.75} \\
\vspace{0.2ex} bert-base-japanese-whole-word-masking & & \\
studio-ousia/mluke-base-lite & 15.00 & 17.00 \\
ku-nlp/deberta-v2-base-japanese & 15.40 & 15.75 \\
cl-tohoku/bert-base-japanese & 16.40 & 18.00 \\
ku-nlp/roberta-base-japanese-char-wwm & 16.60 & 22.25 \\
cl-tohoku/bert-base-japanese-char-v3 & 17.40 & 12.75 \\
cl-tohoku/bert-base-japanese-char-v2 & 19.60 & 15.25 \\
bert-base-multilingual-cased & 20.60 & 20.75 \\
cl-tohoku/bert-base-japanese-char & 22.00 & 19.50 \\
xlm-roberta-base & 22.60 & 22.00 \\
megagonlabs/roberta-long-japanese & 22.80 & 24.00 \\
\bhline
\end{tabular}
\caption{
For each model, we trained on both supervised and unsupervised datasets, and then calculated their average rankings. ``Sup.'' represents the results when fine-tuning with Supervised SimCSE, while ``Unsup.'' indicates the results when using Unsupervised SimCSE.
For example, the column under ``Sup.'' for the \texttt{nlp-waseda/roberta-large-japanese} model represents the average ranking obtained by training the model individually on datasets such as JSICK, JaNLI, NU-NLI, NU-SNLI, and NU-MNLI.
}
\label{tab:model-ranking}

\vspace{4ex}

\tabcolsep 3pt
\begin{tabular}{lcccc@{\hspace{1ex}}c}
\bhline
\tabH Dataset & JSICK (test) & JSTS (train) & JSTS (val) & Avg. \\
\bhline
\tabH JSNLI & 82.63 & 79.44 & 82.98 & \textbf{81.68} \\
JaNLI & 80.92 & 73.41 & 77.98 & 77.44 \\
NU-SNLI & 82.49 & 79.20 & 82.34 & 81.34 \\
NU-MNLI & 75.39 & 81.13 & 83.53 & 80.02 \\
NU-NLI & 81.87 & 79.59 & 82.81 & 81.43 \\
\bhline
\end{tabular}
\caption{
Results when the model is fixed to Tohoku University's BERT\protect\supplement{cl-tohoku/bert-large-japanese-v2} in the context of Supervised SimCSE.
}
\label{tab:sup-dataset-comparison}

\vspace{4ex}

\tabcolsep 3pt
\begin{tabular}{lccccc}
\bhline
\tabH Dataset & JSICK (test) & JSTS (train) & JSTS (val) & Avg. \\
\bhline
\tabH Wiki40B &  79.14 & 77.18 & 81.00 & \textbf{79.11} \\
Wikipedia & 79.40 & 77.18 & 80.28 & 78.95 \\
BCCWJ & 79.45 & 76.71 & 80.83 & 79.00 \\
CC100 & 76.27 & 71.39 & 75.91 & 74.52 \\
\bhline
\end{tabular}
\caption{
Results when the model is fixed to Tohoku University's BERT\protect\supplement{cl-tohoku/bert-large-japanese-v2} in the context of Unsupervised SimCSE.
}
\label{tab:unsup-dataset-comparison}
\end{table}

\begin{table}[t!]
\small
\tabcolsep 7pt
\centering

\begin{tabular}{@{\hspace{2ex}}lc@{\hspace{8ex}}lc@{\hspace{2ex}}}
\bhline
\multicolumn{2}{l}{\tabH \hspace{4ex}Supervised} & \multicolumn{2}{l}{Unsupervised}\\
\tabH Dataset & Rank & Dataset & Rank \\
\bhline
\tabH JSNLI & 1.583 & Wikipedia & 1.875 \\
NU-NLI & 2.000 & BCCWJ & 2.083 \\
NU-SNLI & 2.417 & Wiki40B & 2.208 \\
NU-MNLI & 4.375 & CC100 & 3.833 \\
JaNLI & 4.625 & & \\
\bhline
\end{tabular}
\caption{
For each supervised and unsupervised dataset, models were trained on the respective dataset and their average rankings were computed.
For example, the row for SNLI represents the average rank of a given model across all dataset, specifically for SNLI, averaged over all models.
}
\label{tab:dataset-ranking}

\vspace{3ex}

\tabcolsep 5pt
\begin{tabular}{@{\ \ }rcccc@{\ \ }}
\bhline
\tabH BS \textbackslash\ LR & 1e-5 & 3e-5 & 5e-5 & Avg. \\
\bhline
\multicolumn{5}{c}{\tabH Supervised} \\
\bhline
\tabH 64 \ \ \ \ & 2.750 & 2.700 & 2.742 & 2.731 \\
128 \ \ \ \ & 2.683 & 2.500 & 2.550 & 2.578 \\
256 \ \ \ \ & 2.242 & 2.400 & 2.442 & 2.361 \\
512 \ \ \ \ & 2.325 & 2.400 & 2.267 & 2.331 \\
\bhline
\multicolumn{5}{c}{\tabH Unsupervised} \\
\bhline
\tabH 64 \ \ \ \ & 1.344 & 2.083 & 2.500 & 1.976 \\
128 \ \ \ \ & 1.969 & 1.656 & 1.854 & 1.826 \\
256 \ \ \ \ & 2.990 & 2.792 & 2.375 & 2.719 \\
512 \ \ \ \ & 3.698 & 3.469 & 3.271 & 3.479 \\
\bhline
\end{tabular}
\caption{
Rankings of batch size calculated for all combinations of models and datasets across different learning rates.
``BS'' represents batch size, ``LR'' represents learning rate, and ``Avg.'' represents the average rank of the batch size for each learning rate, respectively.
}
\label{tab:batch-size-ranking}
\end{table}

Table~\ref{tab:main-sup} shows the results of Supervised SimCSE, while Table~\ref{tab:main-unsup} shows those of Unsupervised SimCSE.
For both methods, the results correspond to the models trained with their optimal hyperparameters.
Due to space constraints, we only depict results for Supervised SimCSE using the JSNLI dataset and for Unsupervised SimCSE using the Wiki40B dataset.
Detailed results for all datasets are available on GitHub\footlink{https://github.com/hppRC/simple-simcse-ja/blob/main/results/sup-simcse/all.csv}\footsep\footlink{https://github.com/hppRC/simple-simcse-ja/blob/main/results/unsup-simcse/all.csv}.

For the results of Supervised SimCSE, as base size models, Tohoku University's BERT\supplement{cl-tohoku/bert-base-japanese-v3, cl-tohoku/bert-base-japanese-v2} exhibited high performance.
As large size models, Tohoku University's BERT\supplement{cl-tohoku/bert-large-japanese-v2}, Studio Ousia's LUKE~\cite{LUKE} based Japanese large model\supplement{studio-ousia/luke-japanese-large-lite}, and Waseda University's RoBERTa\supplement{nlp-waseda/roberta-large-japanese} demonstrated superior performance.
Consistent with the results presented in the original SimCSE paper, the large-size models generally outperformed the base-size models. However, the performance gap between the base-size and large-size models was not particularly significant.

For the results of Unsupervised SimCSE, we observed a trend similar to that of Supervised SimCSE where the large-size models generally outperformed the base-size ones.
Notably, the performance of the large-size models, Tohoku University's BERT and Waseda University's RoBERTa, demonstrated superior performance even in the unsupervised setting, surpassing that of the base-size models under the supervised setting.

Overall, we observed a trend in both supervised and unsupervised settings that subword-level models consistently outperformed character-level models.
Regarding multilingual models, mLUKE~\cite{mLUKE} demonstrated consistently better performance compared to multilingual BERT~\cite{BERT} and XLM-RoBERTa~\cite{XLM-RoBERTa}.
For subword-level language models, the choice of tokenizer did not have a significant impact on performance in our setting.

\begin{table*}[t!]
\small
\tabcolsep 5pt
\centering
\begin{tabular}{@{\hspace{2ex}}lcccc@{\hspace{1ex}}}
\bhline
\tabH \hspace{-0.3ex}Model Id & Training Dataset & Learning Rate & Batch Size & STS Avg. \\
\bhline
\tabH \href{https://huggingface.co/cl-nagoya/sup-simcse-ja-large}{cl-nagoya/sup-simcse-ja-large} & JSNLI & 5e-5 & 512 & 81.91 \\
\href{https://huggingface.co/cl-nagoya/sup-simcse-ja-base}{cl-nagoya/sup-simcse-ja-base} & JSNLI & 5e-5 & 512 & 80.49 \\
\href{https://huggingface.co/cl-nagoya/unsup-simcse-ja-large}{cl-nagoya/unsup-simcse-ja-large} & Wiki40B & 3e-5 & 64 & 79.60 \\
\href{https://huggingface.co/cl-nagoya/unsup-simcse-ja-base}{cl-nagoya/unsup-simcse-ja-base} & Wiki40B & 5e-5 & 64 & 77.48 \\
\bhline
\end{tabular}
\caption{
The configurations chosen for training models for public releases, along with the performance of the models.
``STS Avg.'' represents the average performance on JSICK (val), JSTS (train), and JSTS (val).
\label{tab:public-release}
}

\vspace{3ex}

\tabcolsep 3pt
\centering
\begin{tabular}{@{\hspace{2ex}}lc@{\hspace{2ex}}ccc@{\hspace{3ex}}c@{\hspace{1ex}}}
\bhline
\tabH \hspace{-1.3ex}Model Id & JSICK (val) & JSICK (test) & JSTS (train) & JSTS (val) & Avg. \\
    \bhline
    \multicolumn{6}{l}{\tabH Ours} \\
    \bhline
    \tabH \href{https://huggingface.co/cl-nagoya/sup-simcse-ja-large}{cl-nagoya/sup-simcse-ja-large} & 84.36 & \textbf{83.05} & \textbf{79.61} & \textbf{83.07} & \textbf{81.91} \\
    \href{https://huggingface.co/cl-nagoya/sup-simcse-ja-base}{cl-nagoya/sup-simcse-ja-base} & 83.62 & 82.75 & 77.86 & 80.86 & 80.49 \\
    \href{https://huggingface.co/cl-nagoya/unsup-simcse-ja-large}{cl-nagoya/unsup-simcse-ja-large} & 79.89 & 79.62 & 77.77 & 81.40 & 79.60 \\
    \href{https://huggingface.co/cl-nagoya/unsup-simcse-ja-base}{cl-nagoya/unsup-simcse-ja-base} & 79.15 & 79.01 & 74.48 & 78.95 & 77.48 \\
    \bhline
    \multicolumn{6}{l}{\tabH Existing Sentence Embedding Models} \\
    \bhline
    \tabH \href{https://huggingface.co/pkshatech/GLuCoSE-base-ja}{pkshatech/GLuCoSE-base-ja} & 76.36 & 75.70 & 78.58 & 81.76 & 78.68 \\
    \href{https://huggingface.co/pkshatech/simcse-ja-bert-base-clcmlp}{pkshatech/simcse-ja-bert-base-clcmlp} & 74.47 & 73.46 & 78.05 & 80.14 & 77.21 \\
    \href{https://huggingface.co/colorfulscoop/sbert-base-ja}{colorfulscoop/sbert-base-ja} & 67.19 & 65.73 & 74.16 & 74.24 & 71.38 \\
    \href{https://huggingface.co/oshizo/sbert-jsnli-luke-japanese-base-lite}{oshizo/sbert-jsnli-luke-japanese-base-lite} & 72.96 & 72.60 & 77.88 & 81.09 & 77.19 \\
    \hline
    \tabH \href{https://huggingface.co/MU-Kindai/Japanese-SimCSE-BERT-large-sup}{MU-Kindai/Japanese-SimCSE-BERT-large-sup} & 77.06 & 77.48 & 70.83 & 75.83 & 74.71 \\
    \href{https://huggingface.co/MU-Kindai/Japanese-SimCSE-BERT-base-sup}{MU-Kindai/Japanese-SimCSE-BERT-base-sup} & 74.10 & 74.19 & 70.08 & 73.26 & 72.51 \\
    \href{https://huggingface.co/MU-Kindai/Japanese-SimCSE-BERT-large-unsup}{MU-Kindai/Japanese-SimCSE-BERT-large-unsup} & 77.63 & 77.69 & 74.05 & 77.77 & 76.50 \\
    \href{https://huggingface.co/MU-Kindai/Japanese-SimCSE-BERT-base-unsup}{MU-Kindai/Japanese-SimCSE-BERT-base-unsup} & 77.25 & 77.44 & 72.84 & 77.12 & 75.80 \\ 
    \href{https://huggingface.co/MU-Kindai/Japanese-MixCSE-BERT-base}{MU-Kindai/Japanese-MixCSE-BERT-base} & 76.72 & 76.94 & 72.40 & 76.23 & 75.19 \\
    \href{https://huggingface.co/MU-Kindai/Japanese-DiffCSE-BERT-base}{MU-Kindai/Japanese-DiffCSE-BERT-base} & 75.61 & 75.83 & 71.62 & 75.81 & 74.42 \\
    \hline
    \tabH \href{https://huggingface.co/intfloat/multilingual-e5-small}{intfloat/multilingual-e5-small} & 82.01 & 81.38 & 74.48 & 78.92 & 78.26 \\
    \href{https://huggingface.co/intfloat/multilingual-e5-base}{intfloat/multilingual-e5-base} & 81.25 & 80.56 & 76.04 & 79.65 & 78.75 \\
    \href{https://huggingface.co/intfloat/multilingual-e5-large}{intfloat/multilingual-e5-large} & 80.57 & 79.39 & 79.16 & 81.85 & 80.13 \\
    \hline
    \tabH \href{https://huggingface.co/sentence-transformers/LaBSE}{sentence-transformers/LaBSE} & 76.54 & 76.77 & 72.15 & 76.12 & 75.02 \\
    \href{https://huggingface.co/sentence-transformers/stsb-xlm-r-multilingual}{sentence-transformers/stsb-xlm-r-multilingual} & 73.09 & 72.00 & 77.83 & 78.43 & 76.09 \\
    \bhline
    \multicolumn{6}{l}{\tabH Vanilla pre-trained Models} \\
    \bhline
    \tabH \href{https://huggingface.co/cl-tohoku/bert-large-japanese-v2}{cl-tohoku/bert-large-japanese-v2 (Mean)} & 67.06 & 67.15 & 66.72 & 70.68 & 68.18 \\
    \href{https://huggingface.co/studio-ousia/luke-japanese-large-lite}{studio-ousia/luke-japanese-large-lite (Mean)} & 62.23 & 60.90 & 65.41 & 68.02 & 64.78 \\
    \href{https://huggingface.co/studio-ousia/mluke-large-lite}{studio-ousia/mluke-large-lite (Mean)} & 60.15 & 59.12 & 51.91 & 52.55 & 54.53 \\
    \href{https://huggingface.co/cl-tohoku/bert-base-japanese-v3}{cl-tohoku/bert-base-japanese-v3 (Mean)} & 70.91 & 70.29 & 69.37 & 74.09 & 71.25 \\
    \href{https://huggingface.co/cl-tohoku/bert-base-japanese-v2}{cl-tohoku/bert-base-japanese-v2 (Mean)} & 70.49 & 70.06 & 66.12 & 70.66 & 68.95 \\
    \href{https://huggingface.co/cl-tohoku/bert-base-japanese-whole-word-masking}{cl-tohoku/bert-base-japanese-whole-word-masking (Mean)} & 69.57 & 69.17 & 63.20 & 67.37 & 66.58 \\ \hline
    \tabH \href{https://huggingface.co/cl-tohoku/bert-large-japanese-v2}{cl-tohoku/bert-large-japanese-v2 (CLS)} & 46.66 & 47.02 & 54.13 & 57.38 & 52.84 \\
    \href{https://huggingface.co/cl-tohoku/bert-base-japanese-v3}{cl-tohoku/bert-base-japanese-v3 (CLS)} & 51.37 & 51.91 & 58.49 & 62.96 & 57.79 \\
    \bhline
    \multicolumn{6}{l}{\tabH Proprietary Model} \\
    \bhline
    \tabH \href{https://platform.openai.com/docs/api-reference/embeddings}{text-embedding-ada-002} & 79.31 & 78.95 & 74.52 & 79.01 & 77.49 \\
    \bhline
\end{tabular}
\caption{
Evaluation results for both existing models and our proposed model.
Values in the table represent the Spearman’s rank correlation coefficient multiplied by 100.
The meaning of each column aligns with that of Table~\ref{tab:main-sup}.
``Vanilla pre-trained Models'' refers to models that directly apply pooling to the output embeddings of pre-trained language models without any fine-tuning.
The annotations (Mean, CLS) inside the parentheses indicate the Mean Pooling, which averages the output embeddings along the sequence direction, and the CLS Pooling, which takes the embedding corresponding to the head token of the output embeddings, respectively.
\label{tab:comparison}
}
\end{table*}

\subsection{Analysis}

For building future Japanese sentence embedding models, it is crucial to quantitatively analyze which models and datasets are suitable.
To examine the robustness of the models to the datasets, we ranked the models for each dataset and computed their average rankings.
Table~\ref{tab:model-ranking} shows the results.
We can observe that the average rank of Waseda University's RoBERTa is high.
While Waseda's RoBERTa requires tokenization using Juman++, its generally high performance suggests that it is a strong model worth considering.

Next, Table~\ref{tab:sup-dataset-comparison} shows the results for each dataset in a supervised setting, with the model fixed to Tohoku University's BERT\supplement{cl-tohoku/bert-large-japanese-v2}.
In our experiments, JSNLI demonstrated the highest performance.
Interestingly, the performance of the machine-translated NU-MNLI, derived from MNLI, is relatively low.
Moreover, Table~\ref{tab:unsup-dataset-comparison} shows the results for each dataset in a unsupervised setting, with the model fixed to Tohoku University's BERT\supplement{cl-tohoku/bert-large-japanese-v2}.
There seems to be a trend suggesting that selecting Wikipedia-based corpora for fine-tuning with Unsupervised SimCSE might be a preferable choice over Web corpora.
Subsequently, we ranked datasets for each model and computed the average ranks of the datasets.
Table~\ref{tab:dataset-ranking} illustrates that selecting JSNLI is likely a good choice.
JaNLI shows relatively lower results, which is possibly due to its smaller size and its specific dataset design.

Finally, we also computed rankings for the hyperparameters.
Table~\ref{tab:batch-size-ranking} shows the average rankings for each batch size and learning rate.
Specifically, we assigned ranks based on the learning rates for each batch size and calculated their average.
From the table, for supervised settings, it appears advisable to opt for a batch size of 512, while for unsupervised settings, a batch size of around 64 or 128 seems preferable.

\subsection{Publicly Released Models}

Having readily accessible models that can serve as baselines for Japanese sentence embeddings is crucial.
Based on the experimental results of this report, we trained models that can be used as baselines for Japanese SimCSE in both supervised and unsupervised settings, and for both the base and large sizes.
The models we trained were fine-tuned with hyperparameters presented in Table~\ref{tab:public-release}, chosen for their performance and ease of use.
For the base models for fine-tuning, we employed \texttt{\href{https://huggingface.co/cl-tohoku/bert-large-japanese-v2}{cl-tohoku/bert-large-japanese-v2}} and \texttt{\href{https://huggingface.co/cl-tohoku/bert-base-japanese-v3}{cl-tohoku/bert-base-japanese-v3}}.
For both supervised and unsupervised settings, to alleviate randomness, we conducted experiments three times, then, selected models for public release with the highest performance on the development set, i.e., JSICK (train).
The trained models are available on HuggingFace.

Furthermore, Table~\ref{tab:comparison} shows the results of our publicly released models compared to existing Japanese sentence embedding models.
As auxiliary baselines, we also include results from using pre-trained language models directly as sentence embedding models without further fine-tuning.
Among all models, our models trained with Supervised SimCSE demonstrated the highest performance, surpassing existing models, as well as JCSE~\cite{JCSE}, another Japanese SimCSE model.
Notably, our models outperformed strong multi-lingual sentence embedding models such as multilingual E5~\cite{E5} and LaBSE~\cite{LaBSE}.
Furthermore, it is worth noting that the performance of sentence embedding models fine-tuned using Unsupervised SimCSE is on par with existing supervised models.

It is worth noting that these evaluation results are specific to the STS tasks and do not guarantee generalization to other tasks like Dense Passage Retrieval.
In particular, GLuCoSE\supplement{pkshatech/GLuCoSE-base-ja} or multilingual E5 was designed to serve various purposes as a sentence embedding model.
Hence, it is conceivable that performance trends may differ in tasks other than STS.

\section{Conclusion}

In this report, we present extensive experiments on various pre-trained language models, supervised and unsupervised training datasets, and hyperparameters to perform a thorough evaluation of Japanese SimCSE.
Furthermore, utilizing the promising models and datasets obtained from our experimental results, we established a strong baseline for Japanese sentence embedding models.
In the experiments on the Japanese STS task, our model exhibited the highest performance.
We hope that our models serve as a baseline and motivate further advancements in Japanese sentence embedding research.


\begingroup
\renewcommand{\emph}[1]{\textit{#1}}
\bibliography{custom}

\begin{thebibliography}{30}
\expandafter\ifx\csname natexlab\endcsname\relax\def\natexlab#1{#1}\fi

\bibitem[{Agirre et~al.(2015)Agirre, Banea, Cardie, Cer, Diab, Gonzalez-Agirre, Guo, Lopez-Gazpio, Maritxalar, Mihalcea, Rigau, Uria, and Wiebe}]{STS15}
Eneko Agirre, Carmen Banea, Claire Cardie, Daniel Cer, Mona Diab, Aitor Gonzalez-Agirre, Weiwei Guo, I{\~n}igo Lopez-Gazpio, Montse Maritxalar, Rada Mihalcea, German Rigau, Larraitz Uria, and Janyce Wiebe. 2015.
\newblock \href {https://doi.org/10.18653/v1/S15-2045} {{SemEval-2015 Task 2: Semantic Textual Similarity, English, Spanish and Pilot on Interpretability}}.
\newblock In \emph{Proceedings of the 9th International Workshop on Semantic Evaluation ({S}em{E}val)}, pages 252--263.

\bibitem[{Agirre et~al.(2014)Agirre, Banea, Cardie, Cer, Diab, Gonzalez-Agirre, Guo, Mihalcea, Rigau, and Wiebe}]{STS14}
Eneko Agirre, Carmen Banea, Claire Cardie, Daniel Cer, Mona Diab, Aitor Gonzalez-Agirre, Weiwei Guo, Rada Mihalcea, German Rigau, and Janyce Wiebe. 2014.
\newblock \href {https://doi.org/10.3115/v1/S14-2010} {{SemEval-2014 Task 10: Multilingual Semantic Textual Similarity}}.
\newblock In \emph{Proceedings of the 8th International Workshop on Semantic Evaluation ({S}em{E}val)}, pages 81--91.

\bibitem[{Agirre et~al.(2016)Agirre, Banea, Cer, Diab, Gonzalez-Agirre, Mihalcea, Rigau, and Wiebe}]{STS16}
Eneko Agirre, Carmen Banea, Daniel Cer, Mona Diab, Aitor Gonzalez-Agirre, Rada Mihalcea, German Rigau, and Janyce Wiebe. 2016.
\newblock \href {https://doi.org/10.18653/v1/S16-1081} {{SemEval-2016 Task 1: Semantic Textual Similarity, Monolingual and Cross-Lingual Evaluation}}.
\newblock In \emph{Proceedings of the 10th International Workshop on Semantic Evaluation (SemEval)}, pages 497--511.

\bibitem[{Agirre et~al.(2012)Agirre, Cer, Diab, and Gonzalez-Agirre}]{STS12}
Eneko Agirre, Daniel Cer, Mona Diab, and Aitor Gonzalez-Agirre. 2012.
\newblock \href {https://www.aclweb.org/anthology/S12-1051} {{SemEval-2012 Task 6: A Pilot on Semantic Textual Similarity}}.
\newblock In \emph{*{SEM} 2012: The First Joint Conference on Lexical and Computational Semantics {--} Semantic Evaluation ({S}em{E}val)}, pages 385--393.

\bibitem[{Agirre et~al.(2013)Agirre, Cer, Diab, Gonzalez-Agirre, and Guo}]{STS13}
Eneko Agirre, Daniel Cer, Mona Diab, Aitor Gonzalez-Agirre, and Weiwei Guo. 2013.
\newblock \href {https://www.aclweb.org/anthology/S13-1004} {{*SEM 2013 shared task: Semantic Textual Similarity}}.
\newblock In \emph{Second Joint Conference on Lexical and Computational Semantics (*{SEM})}, pages 32--43.

\bibitem[{Akiba et~al.(2019)Akiba, Sano, Yanase, Ohta, and Koyama}]{Optuna}
Takuya Akiba, Shotaro Sano, Toshihiko Yanase, Takeru Ohta, and Masanori Koyama. 2019.
\newblock \href {https://api.semanticscholar.org/CorpusID:196194314} {{Optuna: A Next-generation Hyperparameter Optimization Framework}}.
\newblock \emph{Proceedings of the 25th ACM SIGKDD International Conference on Knowledge Discovery \& Data Mining (KDD)}.

\bibitem[{Bowman et~al.(2015)Bowman, Angeli, Potts, and Manning}]{SNLI}
Samuel~R. Bowman, Gabor Angeli, Christopher Potts, and Christopher~D. Manning. 2015.
\newblock \href {https://doi.org/10.18653/v1/D15-1075} {{A large annotated corpus for learning natural language inference}}.
\newblock In \emph{Proceedings of the 2015 Conference on Empirical Methods in Natural Language Processing (EMNLP)}, pages 632--642.

\bibitem[{Cer et~al.(2017)Cer, Diab, Agirre, Lopez-Gazpio, and Specia}]{STSB}
Daniel Cer, Mona Diab, Eneko Agirre, I{\~n}igo Lopez-Gazpio, and Lucia Specia. 2017.
\newblock \href {https://doi.org/10.18653/v1/S17-2001} {{SemEval-2017 Task 1: Semantic Textual Similarity Multilingual and Crosslingual Focused Evaluation}}.
\newblock In \emph{Proceedings of the 11th International Workshop on Semantic Evaluation ({S}em{E}val)}, pages 1--14.

\bibitem[{Chen et~al.(2023)Chen, Handa, and Shirahama}]{JCSE}
Zihao Chen, Hisashi Handa, and Kimiaki Shirahama. 2023.
\newblock \href {http://arxiv.org/abs/2301.08193} {{JCSE: Contrastive Learning of Japanese Sentence Embeddings and Its Applications}}.
\newblock \emph{arXiv:2301.08193}.

\bibitem[{Chuang et~al.(2022)Chuang, Dangovski, Luo, Zhang, Chang, Soljacic, Li, Yih, Kim, and Glass}]{DiffCSE}
Yung-Sung Chuang, Rumen Dangovski, Hongyin Luo, Yang Zhang, Shiyu Chang, Marin Soljacic, Shang-Wen Li, Scott Yih, Yoon Kim, and James Glass. 2022.
\newblock \href {https://doi.org/10.18653/v1/2022.naacl-main.311} {{DiffCSE: Difference-based Contrastive Learning for Sentence Embeddings}}.
\newblock In \emph{Proceedings of the 2022 Conference of the North American Chapter of the Association for Computational Linguistics: Human Language Technologies (NAACL-HLT)}, pages 4207--4218.

\bibitem[{Conneau et~al.(2020{\natexlab{a}})Conneau, Khandelwal, Goyal, Chaudhary, Wenzek, Guzm{\'a}n, Grave, Ott, Zettlemoyer, and Stoyanov}]{CC100-1}
Alexis Conneau, Kartikay Khandelwal, Naman Goyal, Vishrav Chaudhary, Guillaume Wenzek, Francisco Guzm{\'a}n, Edouard Grave, Myle Ott, Luke Zettlemoyer, and Veselin Stoyanov. 2020{\natexlab{a}}.
\newblock \href {https://doi.org/10.18653/v1/2020.acl-main.747} {{Unsupervised Cross-lingual Representation Learning at Scale}}.
\newblock In \emph{Proceedings of the 58th Annual Meeting of the Association for Computational Linguistics (ACL)}, pages 8440--8451.

\bibitem[{Conneau et~al.(2020{\natexlab{b}})Conneau, Khandelwal, Goyal, Chaudhary, Wenzek, Guzm{\'a}n, Grave, Ott, Zettlemoyer, and Stoyanov}]{XLM-RoBERTa}
Alexis Conneau, Kartikay Khandelwal, Naman Goyal, Vishrav Chaudhary, Guillaume Wenzek, Francisco Guzm{\'a}n, Edouard Grave, Myle Ott, Luke Zettlemoyer, and Veselin Stoyanov. 2020{\natexlab{b}}.
\newblock \href {https://doi.org/10.18653/v1/2020.acl-main.747} {{Unsupervised Cross-lingual Representation Learning at Scale}}.
\newblock In \emph{Proceedings of the 58th Annual Meeting of the Association for Computational Linguistics (ACL)}, pages 8440--8451.

\bibitem[{Devlin et~al.(2019)Devlin, Chang, Lee, and Toutanova}]{BERT}
Jacob Devlin, Ming-Wei Chang, Kenton Lee, and Kristina Toutanova. 2019.
\newblock \href {https://doi.org/10.18653/v1/N19-1423} {{BERT: Pre-training of Deep Bidirectional Transformers for Language Understanding}}.
\newblock In \emph{Proceedings of the 2019 Conference of the North {A}merican Chapter of the Association for Computational Linguistics: Human Language Technologies (NAACL)}, pages 4171--4186.

\bibitem[{Feng et~al.(2022)Feng, Yang, Cer, Arivazhagan, and Wang}]{LaBSE}
Fangxiaoyu Feng, Yinfei Yang, Daniel Cer, Naveen Arivazhagan, and Wei Wang. 2022.
\newblock \href {https://doi.org/10.18653/v1/2022.acl-long.62} {{Language-agnostic BERT Sentence Embedding}}.
\newblock In \emph{Proceedings of the 60th Annual Meeting of the Association for Computational Linguistics (ACL)}, pages 878--891. Association for Computational Linguistics.

\bibitem[{Gao et~al.(2021)Gao, Yao, and Chen}]{SimCSE}
Tianyu Gao, Xingcheng Yao, and Danqi Chen. 2021.
\newblock \href {https://aclanthology.org/2021.emnlp-main.552} {{SimCSE: Simple Contrastive Learning of Sentence Embeddings}}.
\newblock In \emph{Proceedings of the 2021 Conference on Empirical Methods in Natural Language Processing (EMNLP)}, pages 6894--6910.

\bibitem[{Guo et~al.(2020)Guo, Dai, Vrande{\v{c}}i{\'c}, and Al-Rfou}]{Wiki40B}
Mandy Guo, Zihang Dai, Denny Vrande{\v{c}}i{\'c}, and Rami Al-Rfou. 2020.
\newblock \href {https://aclanthology.org/2020.lrec-1.297} {{Wiki-40B: Multilingual Language Model Dataset}}.
\newblock In \emph{Proceedings of the Twelfth Language Resources and Evaluation Conference (LREC)}, pages 2440--2452.

\bibitem[{Jiang et~al.(2023)Jiang, Huang, Luan, Wang, and Zhuang}]{PromptEOL}
Ting Jiang, Shaohan Huang, Zhongzhi Luan, Deqing Wang, and Fuzhen Zhuang. 2023.
\newblock \href {http://arxiv.org/abs/2307.16645} {{Scaling Sentence Embeddings with Large Language Models}}.
\newblock \emph{arXiv:2307.16645}.

\bibitem[{Jiang et~al.(2022)Jiang, Huang, Zhang, Wang, Zhuang, Wei, Huang, Zhang, and Zhang}]{PromptBERT}
Ting Jiang, Shaohan Huang, Zihan Zhang, Deqing Wang, Fuzhen Zhuang, Furu Wei, Haizhen Huang, Liangjie Zhang, and Qi~Zhang. 2022.
\newblock \href {http://arxiv.org/abs/2201.04337} {{PromptBERT: Improving BERT Sentence Embeddings with Prompts}}.
\newblock \emph{arXiv:2201.04337}.

\bibitem[{Kurihara et~al.(2022)Kurihara, Kawahara, and Shibata}]{JGLUE}
Kentaro Kurihara, Daisuke Kawahara, and Tomohide Shibata. 2022.
\newblock \href {https://aclanthology.org/2022.lrec-1.317} {{JGLUE: Japanese General Language Understanding Evaluation}}.
\newblock In \emph{Proceedings of the Thirteenth Language Resources and Evaluation Conference (LREC)}, pages 2957--2966.

\bibitem[{Loshchilov and Hutter(2019)}]{AdamW}
Ilya Loshchilov and Frank Hutter. 2019.
\newblock \href {https://openreview.net/forum?id=Bkg6RiCqY7} {{Decoupled Weight Decay Regularization}}.
\newblock In \emph{International Conference on Learning Representations (ICLR)}.

\bibitem[{Marelli et~al.(2014)Marelli, Menini, Baroni, Bentivogli, Bernardi, and Zamparelli}]{SICK}
Marco Marelli, Stefano Menini, Marco Baroni, Luisa Bentivogli, Raffaella Bernardi, and Roberto Zamparelli. 2014.
\newblock \href {http://www.lrec-conf.org/proceedings/lrec2014/pdf/363_Paper.pdf} {{A SICK cure for the evaluation of compositional distributional semantic models}}.
\newblock In \emph{Proceedings of the Ninth International Conference on Language Resources and Evaluation (LREC)}, pages 216--223.

\bibitem[{Miyazaki and Shimizu(2016)}]{YJCaptions}
Takashi Miyazaki and Nobuyuki Shimizu. 2016.
\newblock \href {https://doi.org/10.18653/v1/P16-1168} {{Cross-Lingual Image Caption Generation}}.
\newblock In \emph{Proceedings of the 54th Annual Meeting of the Association for Computational Linguistics (ACL)}, pages 1780--1790.

\bibitem[{Morita et~al.(2015)Morita, Kawahara, and Kurohashi}]{Juman}
Hajime Morita, Daisuke Kawahara, and Sadao Kurohashi. 2015.
\newblock \href {https://doi.org/10.18653/v1/D15-1276} {{Morphological Analysis for Unsegmented Languages using Recurrent Neural Network Language Model}}.
\newblock In \emph{Proceedings of the 2015 Conference on Empirical Methods in Natural Language Processing (EMNLP)}, pages 2292--2297.

\bibitem[{Ri et~al.(2022)Ri, Yamada, and Tsuruoka}]{mLUKE}
Ryokan Ri, Ikuya Yamada, and Yoshimasa Tsuruoka. 2022.
\newblock \href {https://doi.org/10.18653/v1/2022.acl-long.505} {{mLUKE: The Power of Entity Representations in Multilingual Pretrained Language Models}}.
\newblock In \emph{Proceedings of the 60th Annual Meeting of the Association for Computational Linguistics (ACL)}, pages 7316--7330.

\bibitem[{Wang et~al.(2022)Wang, Yang, Huang, Jiao, Yang, Jiang, Majumder, and Wei}]{E5}
Liang Wang, Nan Yang, Xiaolong Huang, Binxing Jiao, Linjun Yang, Daxin Jiang, Rangan Majumder, and Furu Wei. 2022.
\newblock \href {http://arxiv.org/abs/2212.03533} {{Text Embeddings by Weakly-Supervised Contrastive Pre-training}}.
\newblock \emph{arXiv:2212.03533}.

\bibitem[{Wenzek et~al.(2020)Wenzek, Lachaux, Conneau, Chaudhary, Guzm{\'a}n, Joulin, and Grave}]{CC100-2}
Guillaume Wenzek, Marie-Anne Lachaux, Alexis Conneau, Vishrav Chaudhary, Francisco Guzm{\'a}n, Armand Joulin, and Edouard Grave. 2020.
\newblock \href {https://aclanthology.org/2020.lrec-1.494} {{CCNet: Extracting High Quality Monolingual Datasets from Web Crawl Data}}.
\newblock In \emph{Proceedings of the Twelfth Language Resources and Evaluation Conference (LREC)}, pages 4003--4012.

\bibitem[{Williams et~al.(2018)Williams, Nangia, and Bowman}]{MNLI}
Adina Williams, Nikita Nangia, and Samuel Bowman. 2018.
\newblock \href {https://doi.org/10.18653/v1/N18-1101} {{A Broad-Coverage Challenge Corpus for Sentence Understanding through Inference}}.
\newblock In \emph{Proceedings of the 2018 Conference of the North American Chapter of the Association for Computational Linguistics: Human Language Technologies (NAACL)}, pages 1112--1122.

\bibitem[{Yamada et~al.(2020)Yamada, Asai, Shindo, Takeda, and Matsumoto}]{LUKE}
Ikuya Yamada, Akari Asai, Hiroyuki Shindo, Hideaki Takeda, and Yuji Matsumoto. 2020.
\newblock \href {https://doi.org/10.18653/v1/2020.emnlp-main.523} {{LUKE: Deep Contextualized Entity Representations with Entity-aware Self-attention}}.
\newblock In \emph{Proceedings of the 2020 Conference on Empirical Methods in Natural Language Processing (EMNLP)}, pages 6442--6454.

\bibitem[{Yanaka and Mineshima(2021)}]{JaNLI}
Hitomi Yanaka and Koji Mineshima. 2021.
\newblock \href {https://doi.org/10.18653/v1/2021.blackboxnlp-1.26} {{Assessing the Generalization Capacity of Pre-trained Language Models through {J}apanese Adversarial Natural Language Inference}}.
\newblock In \emph{Proceedings of the Fourth BlackboxNLP Workshop on Analyzing and Interpreting Neural Networks for NLP (BlackboxNLP)}, pages 337--349.

\bibitem[{Yanaka and Mineshima(2022)}]{JSICK}
Hitomi Yanaka and Koji Mineshima. 2022.
\newblock \href {https://doi.org/10.1162/tacl_a_00518} {{Compositional Evaluation on Japanese Textual Entailment and Similarity}}.
\newblock \emph{Transactions of the Association for Computational Linguistics (TACL)}, 10:1266--1284.

\end{thebibliography}
\endgroup

\end{document}